\documentclass[letterpaper,10pt,conference]{ieeeconf}
\IEEEoverridecommandlockouts
\usepackage{import}
\usepackage{times}
\usepackage{multicol}
\usepackage{amsmath,amssymb}
\usepackage{tensor}
\usepackage{graphicx}
\usepackage{subcaption}
\usepackage{overpic}
\usepackage{rotating}
\usepackage{verbatim}
\graphicspath{{./figs/}}
\usepackage{algorithm}
\usepackage{algpseudocode}
\usepackage{color}
\usepackage{soul}
\usepackage{bm}
\usepackage{multirow}
\usepackage{multicol}
\usepackage{arydshln}
\usepackage{xspace}
\usepackage{graphicx}
\usepackage{todonotes}
\usepackage{float}
\usepackage{cite}
\usepackage{booktabs}
\usepackage{xspace}
\usepackage{url}
\usepackage{amssymb}
\usepackage{multirow}
\usepackage{tabularx}

\newcommand{\alg}{\text{POSH}\xspace}

\title{\LARGE \bf Online Motion Planning Over Multiple Homotopy Classes with Gaussian Process Inference}

\author{Keshav Kolur\textsuperscript{*}, Sahit Chintalapudi\textsuperscript{*}, Byron Boots, and Mustafa Mukadam%
\thanks{\textsuperscript{*}Equal contribution.\newline
\indent The authors are with the Robot Learning Lab, Georgia Institute of Technology, USA. Email: \texttt{\{kkolur3, schintalapudi\}@gatech.edu}.\newline
}}

\begin{document}
\maketitle
\thispagestyle{empty}
\pagestyle{empty}


\begin{abstract}
Efficient planning in dynamic and uncertain environments is a fundamental challenge in robotics. In the context of trajectory optimization, the feasibility of paths can change as the environment evolves. Therefore, it can be beneficial to reason about multiple possible paths simultaneously. We build on prior work that considers graph-based trajectories to find solutions in multiple homotopy classes concurrently. Specifically, we extend this previous work to an online setting where the unreachable (in time) part of the graph is pruned and the remaining graph is reoptimized at every time step. As the robot moves within the graph on the path that is most promising, the pruning and reoptimization allows us to retain  candidate paths that may become more viable in the future as the environment changes, essentially enabling the robot to dynamically switch between numerous homotopy classes. We compare our approach against prior work without the homotopy switching capability and show improved performance across several metrics in simulation with a 2D robot in multiple dynamic environments under noisy measurements and execution.
\end{abstract}

\section{Introduction \& Related Work}\label{sec:intro}

Motion planning is a core problem for robotic systems that must successfully navigate their surroundings. In uncertain and dynamic environments, the planning problem becomes especially challenging since a previously feasible solution can quickly become infeasible and must be recomputed. 
Broadly speaking, motion planning approaches can be grouped into the categories of search, sampling, and optimization.

Search-based algorithms, such as A*~\cite{astar} guarantee completeness and optimality, but quickly succumb to the curse of dimensionality as discretization of the environment is necessary.
Sampling-based algorithms, including PRM~\cite{kavraki1996probabilistic} and RRT~\cite{kuffner2000rrt}, are probabilistically complete, generalize better to higher dimensional problems, and have been extended to provide (asymptotic) optimality~\cite{karaman2010incremental}. While these methods are well suited to solve complex problems in environments like 2D mazes, in practice they tend to be very computationally expensive for high dimensional systems and require additional post-processing to smooth the solutions.
Conversely, trajectory optimization algorithms initialize a trajectory from start to goal state and then optimize the path by minimizing some cost function~\cite{zucker2013chomp,schulman2014motion}. GPMP2~\cite{Mukadam-IJRR-18} represents smooth, continuous trajectories as samples from a Gaussian Process (GP) and solves the motion planning problem by performing probabilistic inference on a factor graph.  While these methods are much faster for high dimensional systems, they are sensitive to initialization and can get stuck in poor local optima. In practice, straight-line initializations have been shown to work well and random restarts are performed if that fails~\cite{zucker2013chomp,schulman2014motion}. Initializing with a feasible solution from a sampling-based planner is also an option~\cite{zucker2013chomp}, but this incurs extra computational burden.

\begin{figure}[!t]
    \centering
    \begin{subfigure}[t]{0.48\linewidth}
		\centering
        \includegraphics[trim={95 95 95 95},clip,width=1\linewidth]{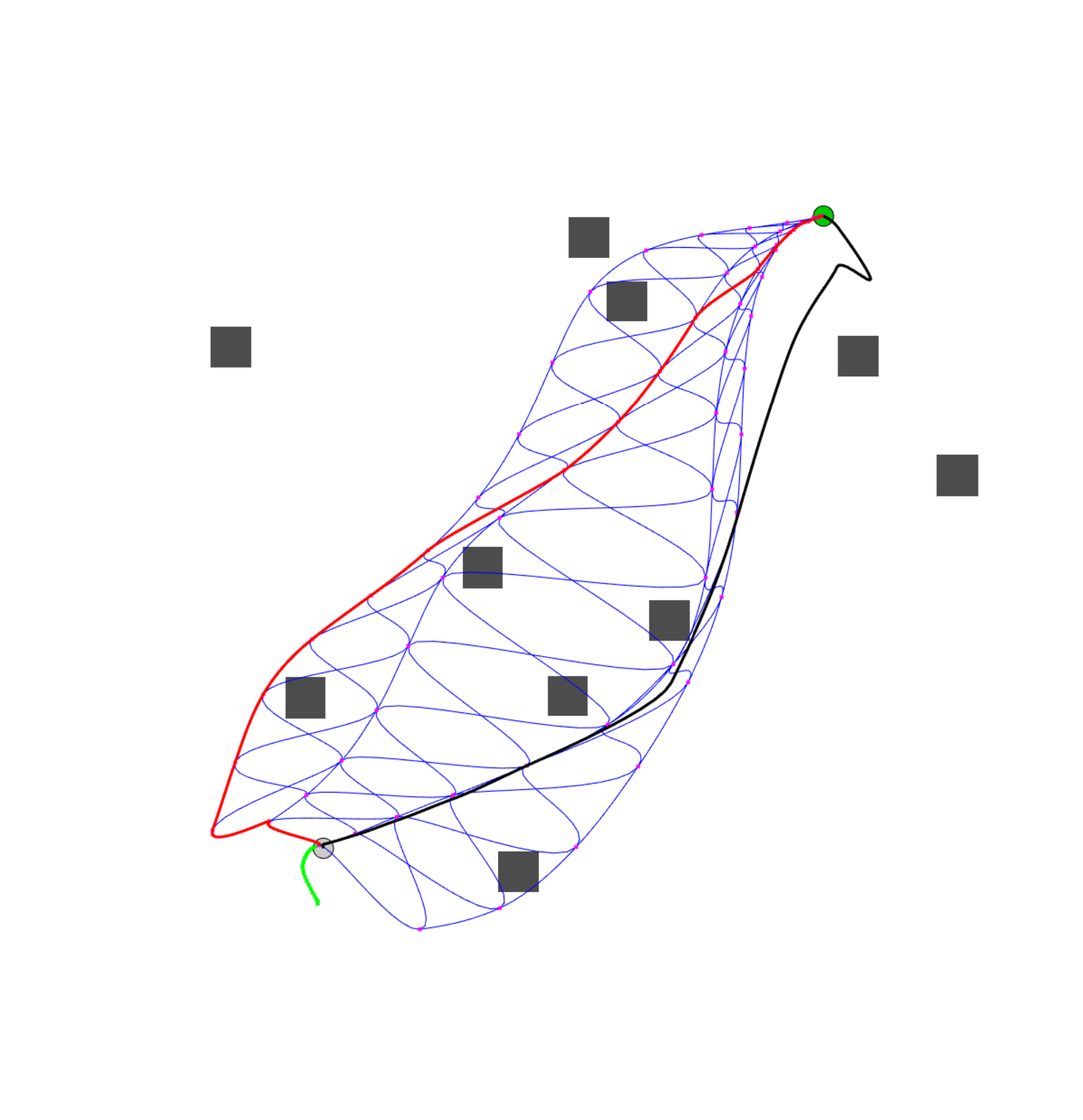}
        \caption{}
		\label{fig:switch1}
    \end{subfigure}
	\hfill\vline\hfill
    \begin{subfigure}[t]{0.48\linewidth}
		\centering
        \includegraphics[trim={95 95 95 95},clip,width=1\linewidth]{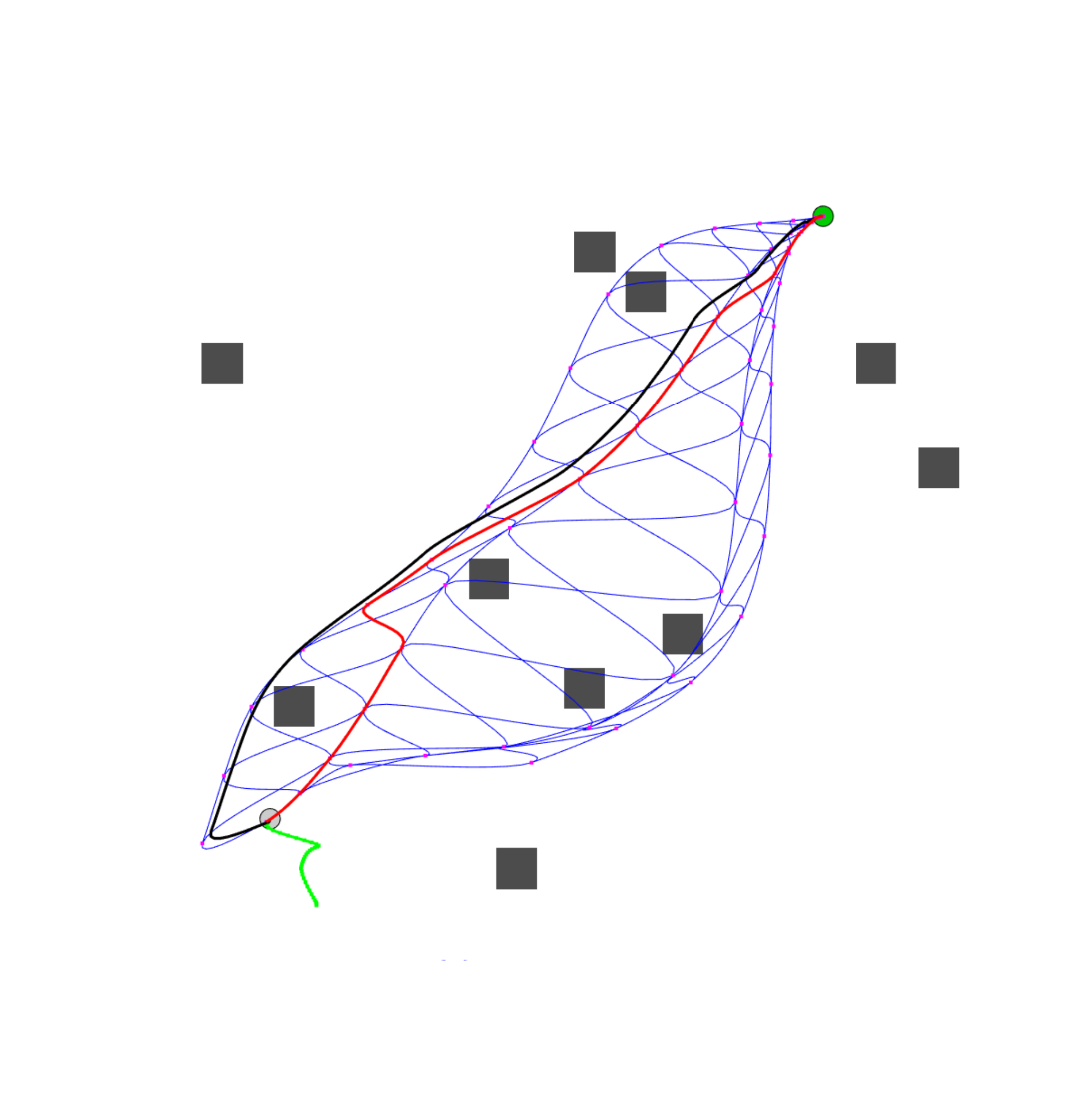}
        \caption{}
		\label{fig:switch2}
    \end{subfigure}
    \caption{\small \alg being used in a dynamic environment where the robot (gray circle) is tasked to reach the goal (green circle) while avoiding obstacles (gray squares). At some time step $i$ In (a) the robot has traveled along the green path when \alg returns a new best (lowest-cost and feasible) trajectory (red). This represents a homotopy switch from the previous best trajectory (black). It executes the first step of this new trajectory and then at the next time step $i+1$ in (b) \alg again returns a new best trajectory (red) that is in a different homotopy class.}
    \label{fig:dynamic_forest}
	\vspace{-3mm}
\end{figure}

Recently, hybrid approaches to planning have attempted to combine the complimentary benefits of different planning schemes. For example, BIT*~\cite{gammell2015batch} combines graph-based and sample-based planning by using a heuristic function to guide a series of random geometric graphs towards the goal state. Its successor, RABIT*~\cite{choudhury2016rabit} performs BIT* while running trajectory optimization when constructing edges between samples, thus reducing sampling complexity by decreasing the number of samples discarded. 
GPMP-GRAPH~\cite{Huang-ICRA-17}, an extension of GPMP2~\cite{Dong-RSS-16}, constructs a graph of interconnected trajectory initializations, similar to the one shown in Fig.~\ref{fig:dynamic_forest}, on a factor graph that when optimized can simultaneously evaluate an exponential number of initializations. After optimization, any path through this graph provides a feasible trajectory. Thus, it has the ability to find solutions in multiple unique homotopy-classes all at once.

While these planning approaches are typically employed in static environments, a number of algorithms~\cite{Park-ITOMP} have been proposed to handle navigating dynamic environments. Lifelong Planning A*~\cite{Koenig-LPA}, D* lite~\cite{SunYK10} and comparable online planning algorithms outperform repeatedly replanning from scratch by incrementally finding shortest paths on a graph with changing edge costs. RRT-X~\cite{Otte2014RRTXRM} an extension of RRT is suited to environments with moving obstacles and provides comparable runtime performance. RAMP~\cite{Vannoy-RAMP} draws from evolutionary computation by maintaining a population of trajectories and evaluating their respective quality (fitness) with respect to a cost function and has been shown to discover and exploit new homotopies. Since solutions can go in and out of feasibility as the environment changes access to solutions in multiple homotopy classes is beneficial. However, there exists a gap in current research with respect to solving the online planning problem by leveraging trajectories across different homotopy classes.

In this work, we present a novel trajectory optimization algorithm, Planning Online by Switching Homotopies (\alg), to handle such scenarios. We build on GPMP-GRAPH~\cite{Huang-ICRA-17} where multiple trajectories are inter-connected and represented as a factor graph upon which probabilistic inference is performed to optimize the entire graph. However, instead of retaining the same optimized trajectory from the initial time step for execution as GPMP-GRAPH does, POSH maintains and updates the entire graph at every time step. Specifically, at any time step the graph is pruned to remove unreachable (in time) states and is then reoptimized considering changes in the environment to find the new optimal trajectory. This grants our algorithm the unique ability to dynamically switch between different homotopy classes as illustrated in Fig.~\ref{fig:dynamic_forest} and allows it to better contend with dynamic environments as demonstrated by our experiments.


\section{Background}\label{sec:back}

In this section we review GPMP-GRAPH, closely following the explanation of Huang et al.~\cite{Huang-ICRA-17}, as well as other concepts, such as homotopy classes and their application to motion planning, necessary to explain our approach. 

\subsection{GPMP-GRAPH}
GPMP-GRAPH is a graph-based trajectory optimization algorithm  designed to address the local minima problem by evaluating an exponential number of initializations simultaneously and, as a consequence, finding solutions in multiple homotopy classes. The graph is constructed as a series of chains between the start and goal state, with connections between the chains. Optimization happens over the entire graph, yielding improvements in planning time over sequentially optimizing a single chain with different initializations~\cite{Huang-ICRA-17}.

GPMP-GRAPH treats the motion planning problem as probabilistic inference. Given desired events $e$, it finds the maximum a posteriori (MAP) trajectory $x^*$
$$x^* = \arg\max_x p(x|e) = \arg\max_x p(x)l(x;e)$$
where $p(x)$ is the prior distribution over trajectories and $l(x;e)$ is the likelihood.

Continuous-time trajectories are represented as samples from a vector-valued Gaussian process (GP) $x(t)\sim \mathcal{GP}(\mu(t), \mathcal{K}(t, t'))$ where $\mu(t)$ is a vector-valued mean function and $\mathcal{K}(t, t')$ is a matrix-valued covariance function. 
Following the definition of GP, trajectories can be parameterized at discrete times $t = \{t_0, \dots, t_n\}$ to be jointly Gaussian distributed
$$x = [x_0, \dots, x_n]^T\sim\mathcal{N}(\mu, \mathcal{K})$$
where:
$$\mu = [\mu(t_0), \dots, \mu(t_n)], \qquad \mathcal{K} = [\mathcal{K}(t_i, t_j)]_{i, j,  0 \leq i,  j \leq n}$$
The values $x_0, \dots, x_n$ are referred to as \textit{support states} and correspond to $x(0),\dots, x(n)$. This defines a prior distribution over trajectories
$$p(x) \propto \exp\Bigg\{-\frac12||x - \mu||^2_\mathcal{K}\Bigg\}$$
In practice, this distribution can be constructed from robot dynamics, for example, modeled as a linear time-varying stochastic differential equation (LTV-SDE)
$$\dot{x}(t) = A(t)x(t) + u(t) + F(t)w(t)$$
where $A$, $F$ are system matrices, $u$ is a known control input, and
$w(t) \propto \mathcal{GP}(0, Q_c\delta(t - t'))$ is a white noise process with $Q_c$ being the power spectral density matrix of the system and $\delta$ being Dirac delta function. 
The first- and second-order moments of the solution to this SDE yield the mean and covariance of the GP, respectively. The prior penalizes the trajectory from deviations from the mean to maintain smoothness in the solutions. The likelihood $l(x;e)$, handles costs like obstacle avoidance and any other motion constraints.

\begin{figure}
    \centering
    \includegraphics[width=0.48\textwidth]{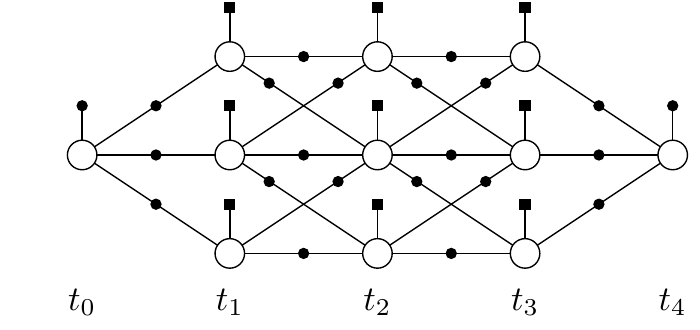}
    \caption{\small Factor Graph consisting of three Gauss-Markov Chains as illustrated in~\cite{Huang-ICRA-17}. The white circles represent support states. Black squares are collision factors and black circles are GP priors factors. Interpolated collision factors are omitted for clarity.}
    \label{fig:factor_graph}
	\vspace{-3mm}
\end{figure}

The posterior distribution $p(x|e)$ can be represented as a \textit{factor graph}. A factor graph is a bipartite graph $G = \{\Theta, F, E\}$, where $\Theta$ is a set of variables, $F$ is a set of factors, and $E$ is the set of edges connecting $F$ to $\Theta$. Then the posterior can be factorized as
$$p(x|e) = \prod_{t_i}f^{gp}_i(x_i, x_{i + 1})f^{obs}_i(x_i)\prod_{\tau}f^{intp}_{i, \tau} (x_i, x_{i + 1})$$
where $f^{gp}_i$ is a binary factor connecting consecutive states, $f_i^{obs}$ is a unary collision likelihood factor, and $f^{intp}_{i, \tau}$ is a collision likelihood factor for a state at $\tau$ between consecutive support states.
This factor graph has a sparse structure which can be exploited for rapidly solving the optimization problem~\cite{Mukadam-IJRR-18}.

So far this is identical to GPMP2~\cite{Mukadam-IJRR-18}, however GPMP-GRAPH constructs such a factor graph with multiple initial trajectories (chains), each with its own set of support states. Adjacent chains have interconnected support states with a GP prior factor and interpolated factors between two states from two different chains, allowing the robot to switch between chains. An example graph construction with three chains and interconnections is shown in Fig.~\ref{fig:factor_graph}. The graph's variables (corresponding to trajectory waypoints) are optimized using the Levenberg-Marquardt algorithm to maximize their probability given the desired events, such as being collision free and conforming to motion constraints. Then, from the optimized graph we can find the trajectory with the lowest cost or the trajectory that conforms to any predefined homotopy constraints set by the user. For more details please refer to~\cite{Huang-ICRA-17}.

GPMP-GRAPH suffers in practical performance because it is unable to exploit the different homotopy classes it identifies when the environment changes. By constructing a factor graph with multiple chains and then allowing those chains to converge to low-cost solutions in different homotopy-classes, the algorithm has many feasible trajectories available to it. However, dynamic obstacles may cause an initially good trajectory to decrease in quality if the obstacle moves in the way of the trajectory. Only optimizing the chain that the robot is on at that time may not allow the robot to switch homotopies, leading to collisions with the moving obstacle or inefficient solutions. 

\subsection{Homotopy Classes}

GPMP-GRAPH improves upon its predecessor GPMP2, by initially considering multiple homotopy classes when optimizing over different initial trajectories in parallel.
Two continuous functions over the real numbers $x_1, x_2:~\mathbb{R}~\rightarrow~\mathbb{R}$ are said to be homotopically equivalent if there exists a continuous function 
$$h: \mathbb{R} \times [0, 1] \rightarrow \mathbb{R}$$
such that $h(t, 0) = x_1(t)$ and $h(t, 1) = x_2(t), t \in \mathbb{R}$.
In the context of motion planning, it is said that two trajectories (with a fixed start and end state) belong to the same homotopy class if one trajectory can be \emph{continuously deformed} into the other without intersecting an obstacle~\cite{Bhattacharya-2012}. This deformation is given by the homotopy function $h$.

We use the concept of \textit{h-signature} to identify the frequency at which the robot changes the homotopy class of its trajectory. An h-signature is a unique identifier for a homotopy class in an environment. Two trajectories have the same h-signature if and only if they belong to the same homotopy class~\cite{Grigoriev-98}. The h-signature for a trajectory is determined by extending a vertical ray upwards towards \textit{y = +$\infty$} from the center of each obstacle and assigning a letter $t_k$ to obstacle $k$. We iterate through each point in the trajectory $x$ from start to goal and check at every time step $t$ if the trajectory crosses a ray between $t$ and $t - 1$. If the trajectory crosses obstacle $k$ from left to right we append $t_k$ to the h-signature of the trajectory and if the trajectory crosses obstacle $k$ from right to left we append $\bar{t_k}$ to the trajectory. The final h-signature is produced by reducing this signature by removing all instances of $t_k\bar{t_k}$ or $\bar{t_k}t_k$ in the signature. \cite{Grigoriev-98} proves that this procedure yields a signature that uniquely corresponds to a homotopy class in the environment. 

By identifying trajectories in different homotopy classes, \alg is finding solutions to the problem that do not deform into each other. This is useful because trajectory-optimization methods like GPMP-GRAPH typically will not deform a solution through an obstacle, preferring to get caught in a local minima in the current homotopy class. By designing \alg such that trajectories in multiple homotopy-classes are maintained, the planner will better be able to adapt to changes in the environment that render the current trajectory suboptimal.






\section{Planning Online by Switching Homotopies}
We begin by motivating \alg with our key insight that switching homotopy classes is essential to navigating changing environments. Then we present our algorithm and its implementation details. 

\subsection{Motivation}
In Fig.~\ref{fig:narrow_passage} we see GPMP-GRAPH try to navigate through a 2-Dimensional environment with a moving obstacle. While at time step $1$ the robot seems to have found a high quality trajectory, changes in the environment start to degrade that trajectory. Namely, reoptimizing a single chain means that the robot will consider only its current homotopy-class, as deforming the trajectory through an obstacle would lead to an increase in trajectory cost. As a consequence, the trajectory is pushed further into a corner as the obstacle approaches, leading to multiple collisions in the trajectory. Intuitively, it seems reasonable to travel along the left of the moving obstacle if it is moving closer, however such a solution would be in a different homotopy class. For this reason, we should maintain and reoptimize trajectories in both the homotopy classes, the one to the right of the obstacle and the one to the left of the obstacle. Then, at every time step the robot can replan a solution to the goal state using the reoptimized chains of the factor graph. As the obstacle moves closer, the low-quality trajectory to the right of the obstacle will suffer a higher cost than a trajectory that moves to the left of the obstacle. Thus, the robot will now respond to this change in trajectory quality by changing the homotopy class mid-execution to avoid collision. 

\subsection{Algorithm}

\begin{figure*}[ht]
    \centering
    \begin{subfigure}[t]{0.241\linewidth}
        \includegraphics[trim={50 80 50 100},clip,width=\linewidth]{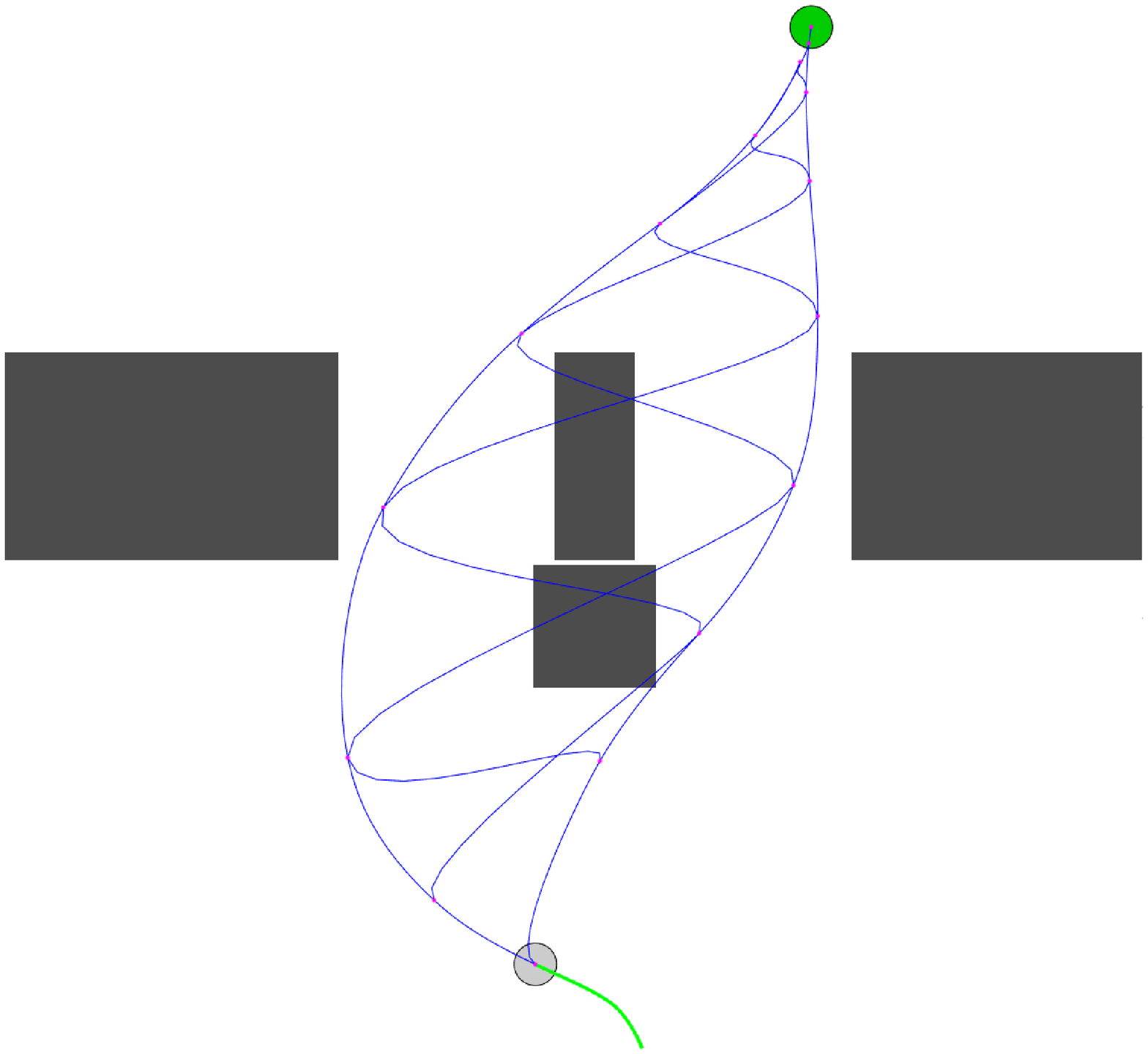}
        \caption{}\label{fig:unoptimized}
    \end{subfigure}%
	\hfill\vline\hfill
    \begin{subfigure}[t]{0.24\linewidth}
        \includegraphics[trim={50 70 50 100},clip,width=\linewidth]{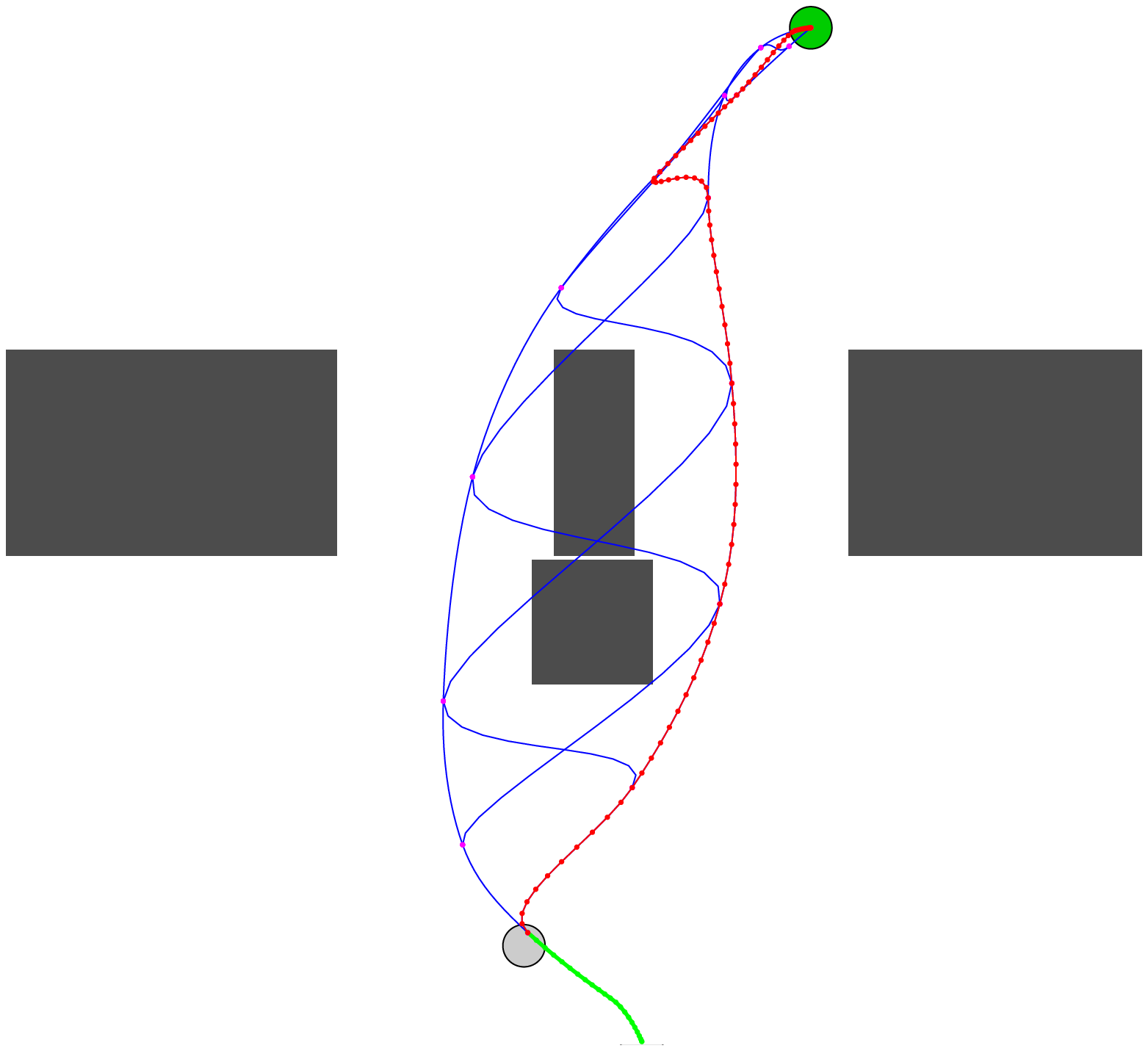}
        \caption{}\label{fig:optimized}
    \end{subfigure}%
	\hfill\vline\hfill
    \begin{subfigure}[t]{0.24\linewidth}
        \includegraphics[trim={50 70 50 100},clip,width=\linewidth]{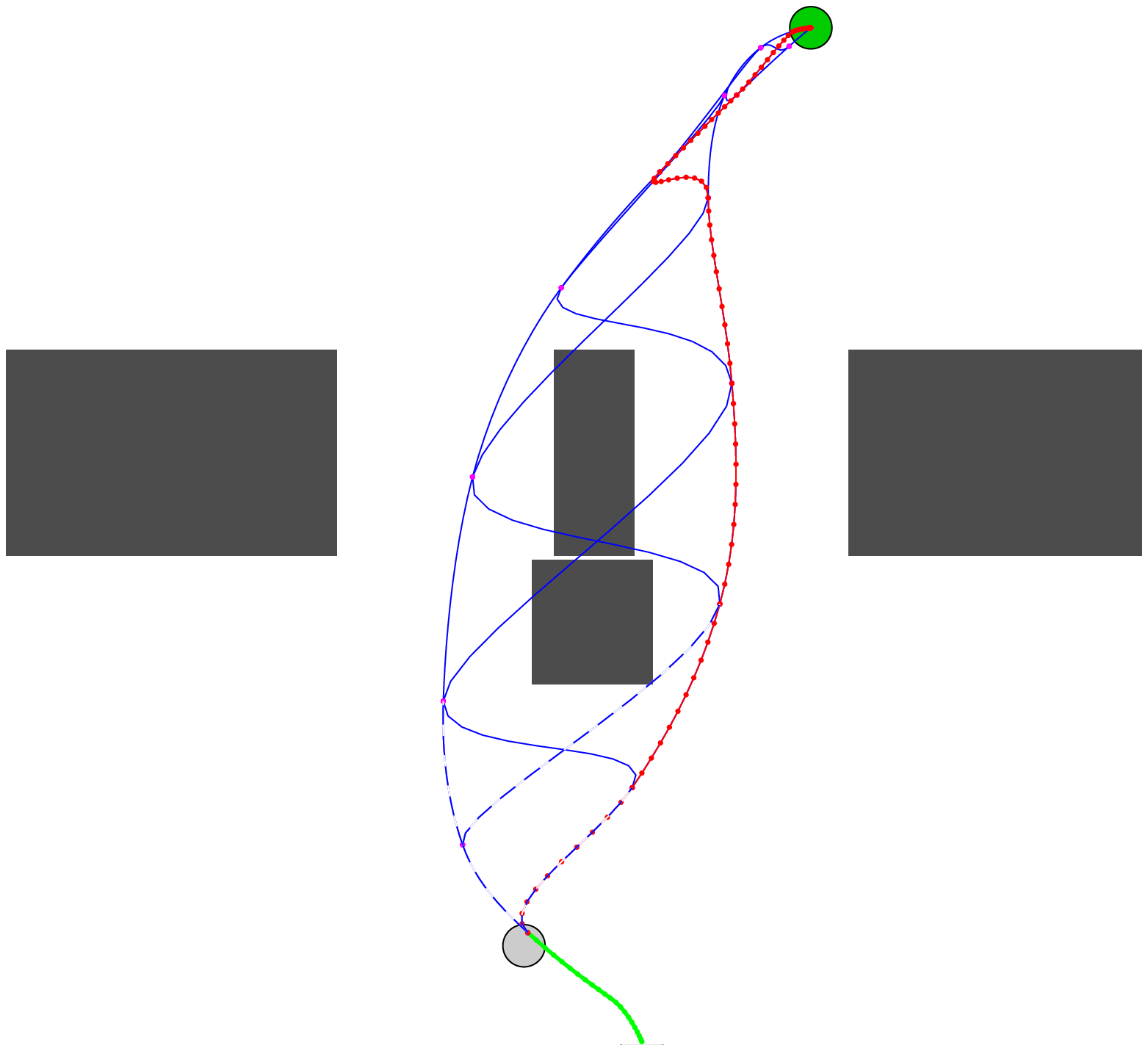}
        \caption{}\label{fig:prune}
    \end{subfigure}%
	\hfill\vline\hfill
    \begin{subfigure}[t]{0.24\linewidth}
        \includegraphics[trim={50 70 50 100},clip,width=\linewidth]{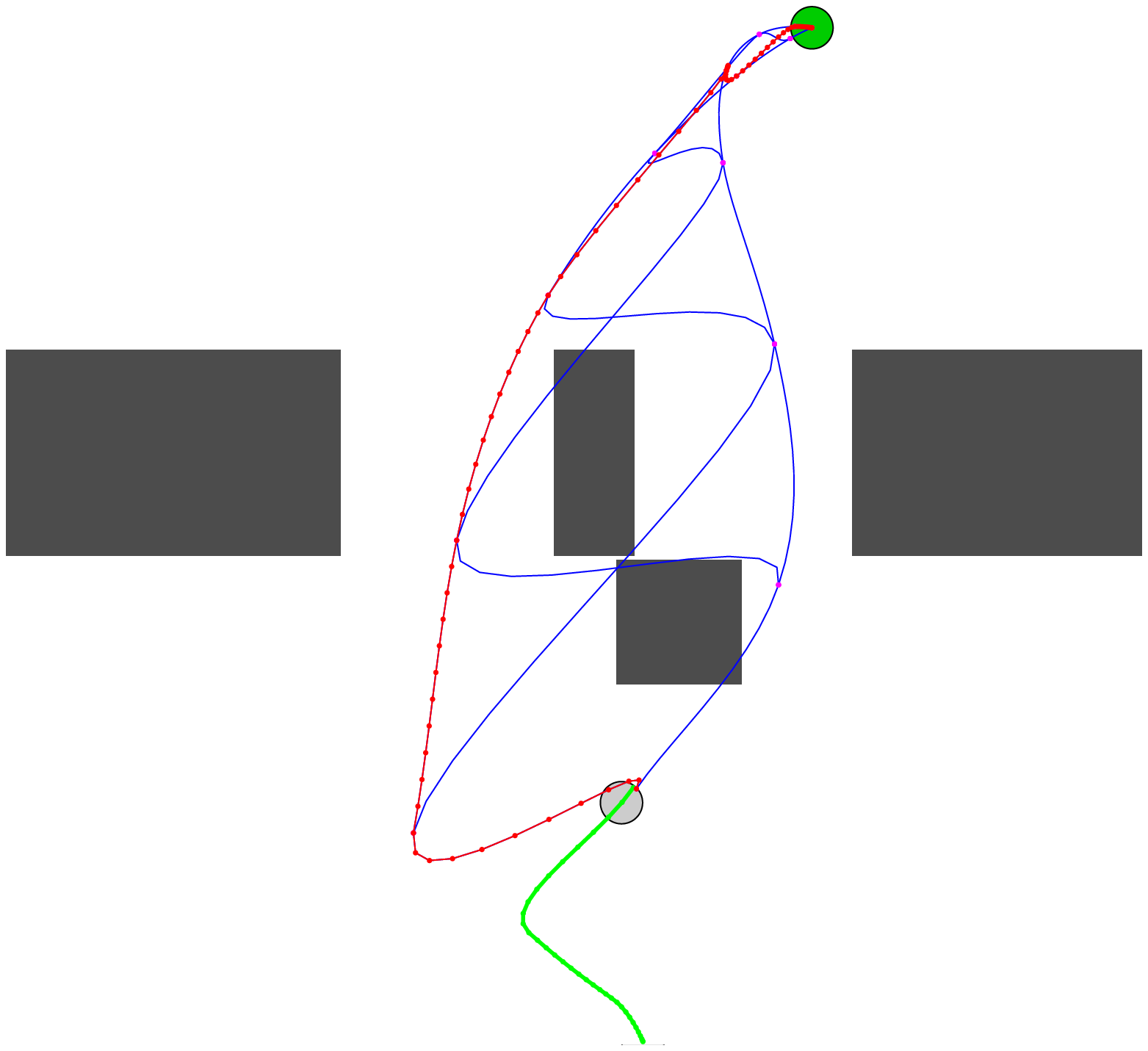}
        \caption{}\label{fig:new_htpy}
    \end{subfigure}
    \caption{\small In (a), the gray circle and green circle represent the robot and goal state respectively. The graph is warm-started from its previous state. In~(b), the graph is reoptimized and then A* is used to find a low-cost trajectory (red) on the graph. In~(c), edges that will no longer be traversable (dashed blue lines) after executing the first control are pruned. Finally in~(d), the robot executes its first control as the environment changes. We see that \alg finds a trajectory in a new homotopy class (red) after reoptimizing.}
    \label{fig:toy_breakdown}
	\vspace{-2mm}
\end{figure*}

\begin{figure}[!t]
	\centering
    \begin{subfigure}[t]{0.32\linewidth}
        \includegraphics[trim={380 50 330 100},clip,width=\linewidth]{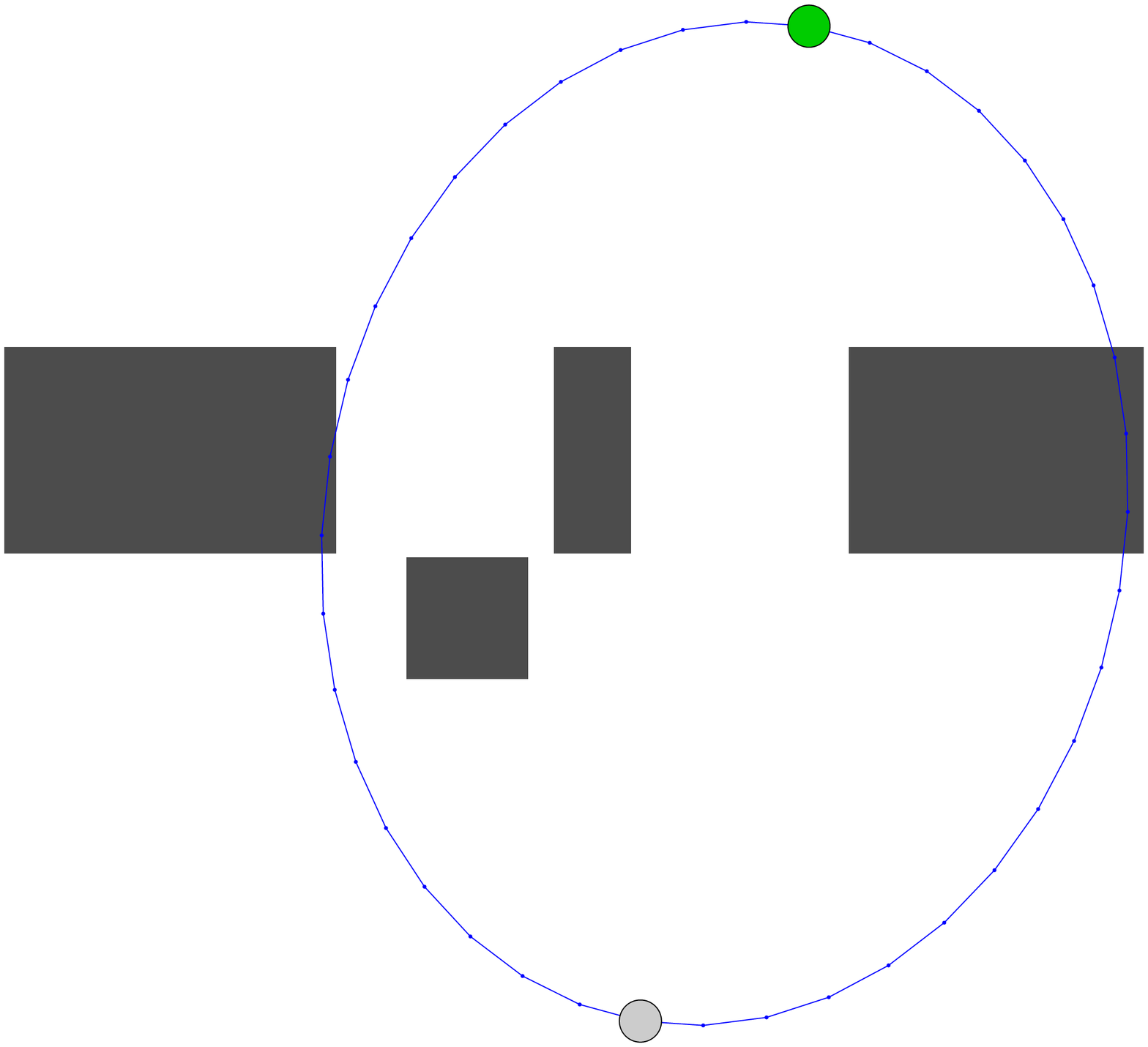}
        \caption{}\label{fig:nointerconnection}
    \end{subfigure}%
	\hfill\vline\hfill
    \begin{subfigure}[t]{0.32\linewidth}
        \includegraphics[trim={380 50 330 100},clip,width=\linewidth]{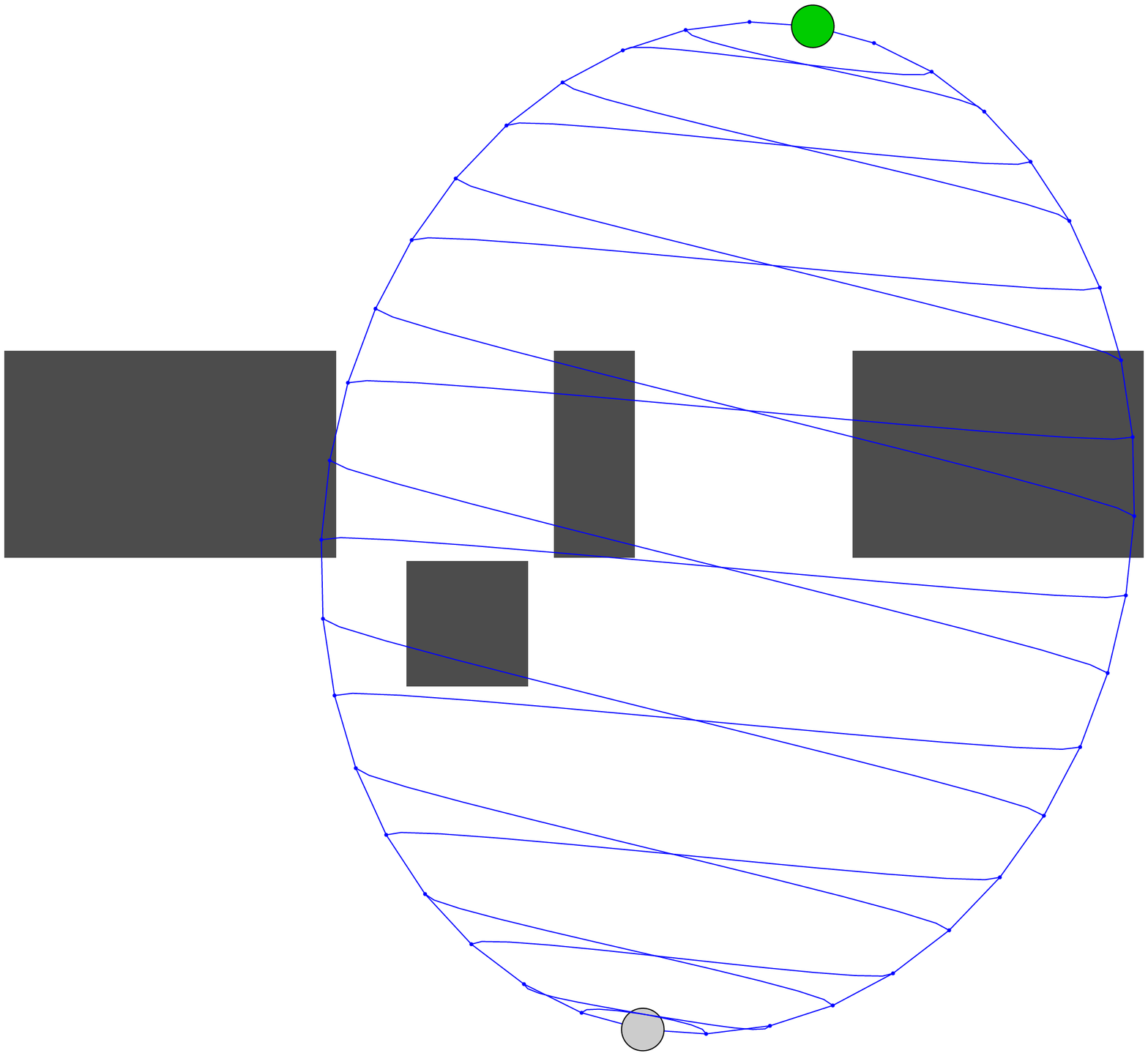}
        \caption{}\label{fig:halfinterconnection}
    \end{subfigure}%
	\hfill\vline\hfill
    \begin{subfigure}[t]{0.32\linewidth}
        \includegraphics[trim={380 50 330 100},clip,width=\linewidth]{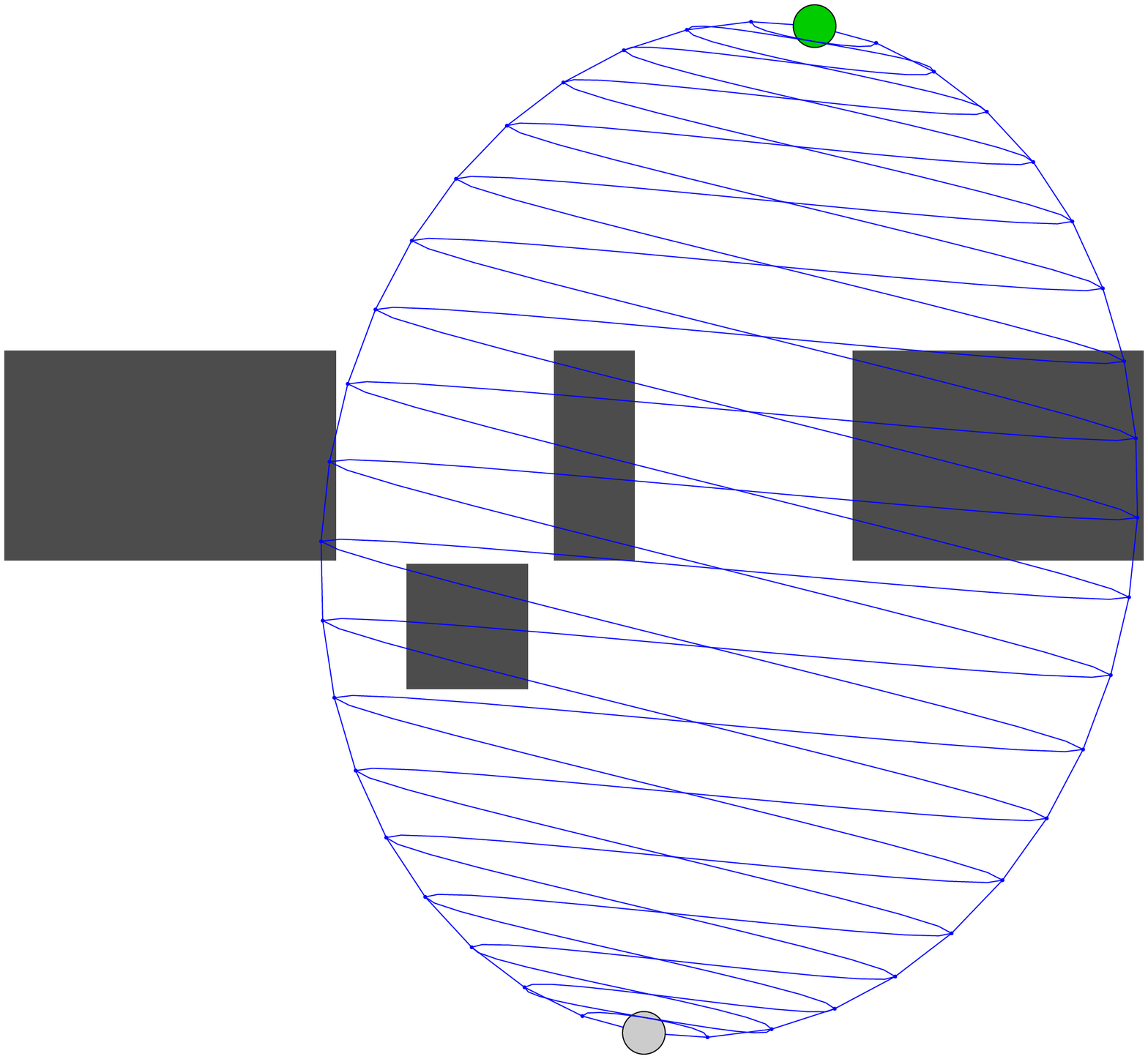}
        \caption{}\label{fig:interconnection}
    \end{subfigure}%
    \caption{\small Possible graph initializations with~(a) no connection between chains,~(b) connections between chains at every other state, and~(c) fully connected chains. The robot is the gray circle and the goal state is the green circle.}
	\vspace{-3mm}
\end{figure}

\alg is summarized by Algorithm 1. The initial obstacle positions are obtained to create the initial SDF as shown by line 1. In line 2, a factor graph is constructed given initial robot states, a signed distance field (SDF) of the environment, and parameters like smoothness prior (desired trade-off between smoothness and collision avoidance) and ratio of interconnections (ranges from 0 i.e. no connections to 1 i.e. every state is interconnected). Given a number of desired trajectories (chains) to be interconnected and a number of time steps, the locations of the initial support states are computed by constructing a series of ellipses with the major radius stretching from the start state to the goal state and varying minor radii as shown in Fig.~\ref{fig:nointerconnection}. Based on the desired ratio of interconnections, states in spatially nearby chains are then interconnected to allow switching between them (see Fig.~\ref{fig:halfinterconnection} and Fig.~\ref{fig:interconnection}). In line 4, given the initial graph and SDF, \alg employs the Levenberg-Marquardt algorithm to optimize the factor graph. 

\begin{algorithm}[!t]
\caption{POSH}\label{alg:plosh}
\begin{algorithmic}[1]
\State $SDF\gets initial\_obstacle\_poses$
\State $factor\_graph \gets \{parameters, robot\_states, SDF\}$
\For{$t = 1, \dots, T$}
    \State $factor\_graph.optimize()$
    \State $x \gets A^*(factor\_graph)$
    \State $factor\_graph.prune\_unreachable()$
    \State $execute\_step(x)$
    \State $SDF.update(obstacle\_poses)$
\EndFor
\end{algorithmic}
\end{algorithm}

Up to this step, \alg has executed the same steps as GPMP-GRAPH would in solving the motion planning problem. Following this, GPMP-GRAPH simply executes the steps in the most feasible trajectory initially obtained and re-optimize this same trajectory at every time step until the goal configuration is reached. However, as discussed earlier this is insufficient to successfully plan in a stochastic, dynamic environment. POSH reoptimizes the entire graph with an updated SDF and re-calculates the most feasible trajectory. These steps, shown in lines 4-8 in Algorithm 1, are executed at every time step until the goal state is reached.
 
We use the example in Fig.~\ref{fig:toy_breakdown} to illustrates the process \alg follows. Fig.~\ref{fig:unoptimized} shows the graph before an iteration of optimization. The robot is the gray circle in the bottom left and the green circle in the top right is the goal state. The graph's current state represents the optimal graph configuration found given the SDF at the last time step. In Fig.~\ref{fig:optimized}, \alg re-optimizes the factor graph based on the updated SDF. Then, as shown in line 5 of Algorithm 1, A* is run on this updated, optimized factor graph to return the lowest cost trajectory (plotted in red). Because A* is complete and optimal, and the graph connects the start and goal states, it is guaranteed that the returned solution will be the lowest cost solution out of possible feasible trajectories encoded in the graph.

This is followed by line 6, in which \alg runs depth first search (DFS) to prune away a subgraph of the factor graph to only maintain the reachable states. Given the current state as the root and the goal state, DFS determines which support states are unreachable in time and are therefore irrelevant. In Fig.~\ref{fig:prune}, the pruned away subgraph is shown with dashed blue lines. By executing the first step (line 7 of Algorithm 1) in the trajectory returned by A* in Fig.~\ref{fig:prune}, the robot arrives at its new position depicted in Fig.~\ref{fig:new_htpy}. At the next time step, the obstacle moves closer to the trajectory the robot was planning to take as Fig.~\ref{fig:new_htpy} shows. Then in line 8 the SDF is updated given changes in the environment. After running another iteration of optimization and A*, the robot now identifies a new trajectory in a different homotopy class that is lower cost and feasible. By doing so, \alg avoids the local minima problem that GPMP-GRAPH is unable to contend with as seen in Fig.~\ref{fig:narrow_passage}, resulting in a collision. On the other hand, \alg successfully adapts to the dynamic environment by switching homotopies. This process repeats until the robot has reached the goal.

\subsection{Implementation Details}

\alg is implemented by building on the GTSAM~\cite{frank2012factor} and GPMP2 library. The factor graph is optimized using the Levenberg-Marquardt algorithm initialized with $\lambda = 10^{-5}$. In order to successfully plan in a stochastic, dynamic environment, \alg must optimize the factor graph based on an updated SDF and determine the most feasible trajectory. Only the first step is executed as this trajectory might not be feasible for future configurations of the environment. To replan efficiently at the next time step, the relevant portions of the optimized factor graph must be retained so that the graph is not reconstructed from scratch. \alg accomplishes this by generating a spanning tree of the graph rooted at the first state in the most feasible trajectory, before navigating to it. The subgraph of the factor graph that does not belong to this spanning tree is pruned away as the states in this subgraph become unreachable in the next time step. Additionally, to replan quicker, the next optimization of the graph is warm-started with the state of the previously initialized factor graph. This results in Levenberg-Marquardt taking less time to optimize the factor graph, since the initialization of the graph was a local minima of the cost function at the last time step, assuming the environment did not drastically change between time steps. 

\section{Results}\label{sec:results}

We benchmark \alg against GPMP2 and GPMP-GRAPH, two trajectory optimization methods within the same family on multiple 2D dynamic environments\footnote{A video of the experiments is available at \newline \indent \url{https://youtu.be/FtUI9VR3iWI}}. GPMP2 and GPMP-GRAPH are set up as a single chain and interconnected multi-chain graphs respectively. At each time step the GPMP2 chain is optimized and a step is taken on the optimized path. Then, the past state is pruned, the noisy state measurement is added, and the chain is reoptimized for the next time step. Since GPMP-GRAPH is an offline batch method, to keep comparisons fair, we will optimize the full multi-chain graph at the first time step to get an optimal sequence of states (a chain), while the other states are pruned away. From then on it is treated as a GPMP2 chain and is optimized accordingly at every time step. In contrast, \alg will continue to keep a pruned graph after every time step to be reoptimized.

These algorithms' performance are tested on two simulated 2D datasets, a dynamic forest and narrow passageway environment. In both environments, the holonomic robot, starting from some initial state, is tasked with reaching the goal while avoiding obstacles. The robot is also subjected to stochastic executions and noisy measurements i.e. the robot's position at every time step is perturbed to simulate execution noise and localization noise is injected into state measurements respectively.
Sections IV-A and IV-B will describe the narrow passageway and dynamic forest environments respectively as well as analyze the algorithms' performance in each. We record the following metrics averaged across multiple runs in each environment: (i) success rate to quantify the runs where the robot gets to the goal without any collisions, (ii) collision intensity to quantify the percentage of the trajectory spent in collisions when they occur, (iii) total distance traveled indicative of smoothness in the path, and (iv) total number of homotopy switches made from reoptimizing at each time step.

For all algorithms, the safety distance, $\epsilon$, and obstacle cost weight, $\sigma_{obs}$ (represents the trade-off between a trajectory smoothness and collision avoidance), are kept constant. The execution prior (initial state) and localization noise levels are all kept constant as well. To account for stochasticity, each algorithm's performance was averaged over 10 Monte-Carlo runs for each experiment. One narrow passageway environment and 13 dynamic forest environments were used to test the algorithms' performance. In the case of GPMP-GRAPH and \alg, hyperparameters like, the number of chains $N_I$, the ratio of interconnections between chains $R_I$, and GP prior noise applied to the interconnections between chains $Q_I$, are tuned to give good performance for the respective algorithm.

\subsection{Dynamic Narrow Passageway Benchmark}

\begin{table}[!t]
	\caption{Benchmark results on narrow passage dataset.}
	\label{table:NP-methodsResults}
	\begin{center}
		\begin{tabular}[c]{rccc}
			\toprule
			Metric & \alg & GPMP-GRAPH & GPMP2 \\
			\midrule
			Success Rate (\%) & \textbf{100} & 10 & 10\\
			Collision Intensity (\%) & \textbf{0} & 6.5 & 7.5\\
			Distance (m) & \textbf{26.7} & 31.6 & 33.8\\
			Homotopy Switches & \textbf{1.2} & 0.2 & 1.1\\
			\bottomrule
		\end{tabular}
	\end{center}
	\vspace{-3mm}
\end{table}

\begin{figure*}[ht]
    \centering
    \begin{subfigure}[t]{0.24\linewidth}
        \includegraphics[trim={60 80 100 90},clip,width=\linewidth]{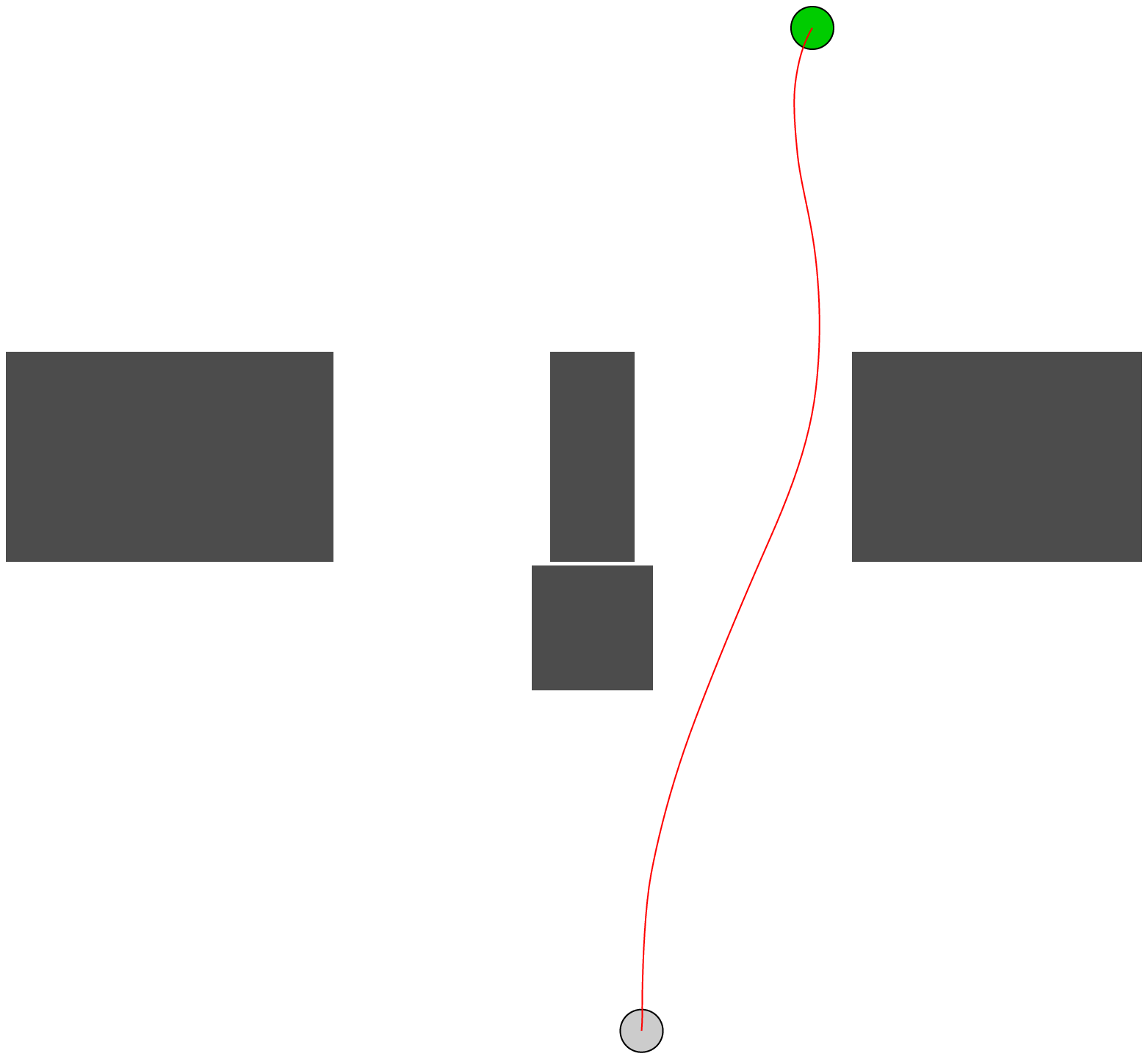}
        \caption{$t=1$}\label{fig:t1}
    \end{subfigure}%
	\hfill\vline\hfill
    \begin{subfigure}[t]{0.24\linewidth}
        \includegraphics[trim={60 80 100 90},clip,width=\linewidth]{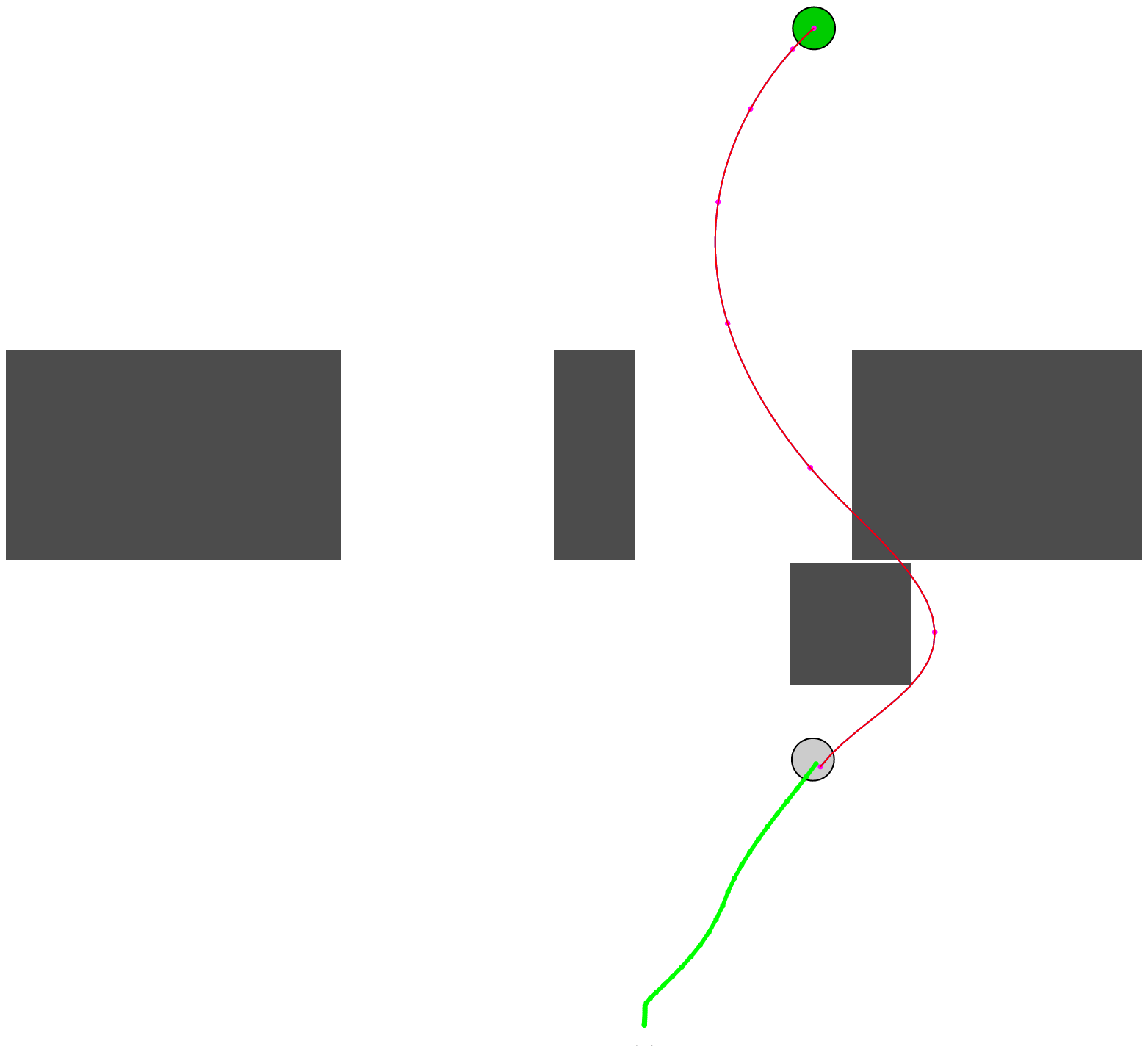}
        \caption{$t=4$}\label{fig:t6}
    \end{subfigure}%
	\hfill\vline\hfill
    \begin{subfigure}[t]{0.24\linewidth}
        \includegraphics[trim={60 80 100 90},clip,width=\linewidth]{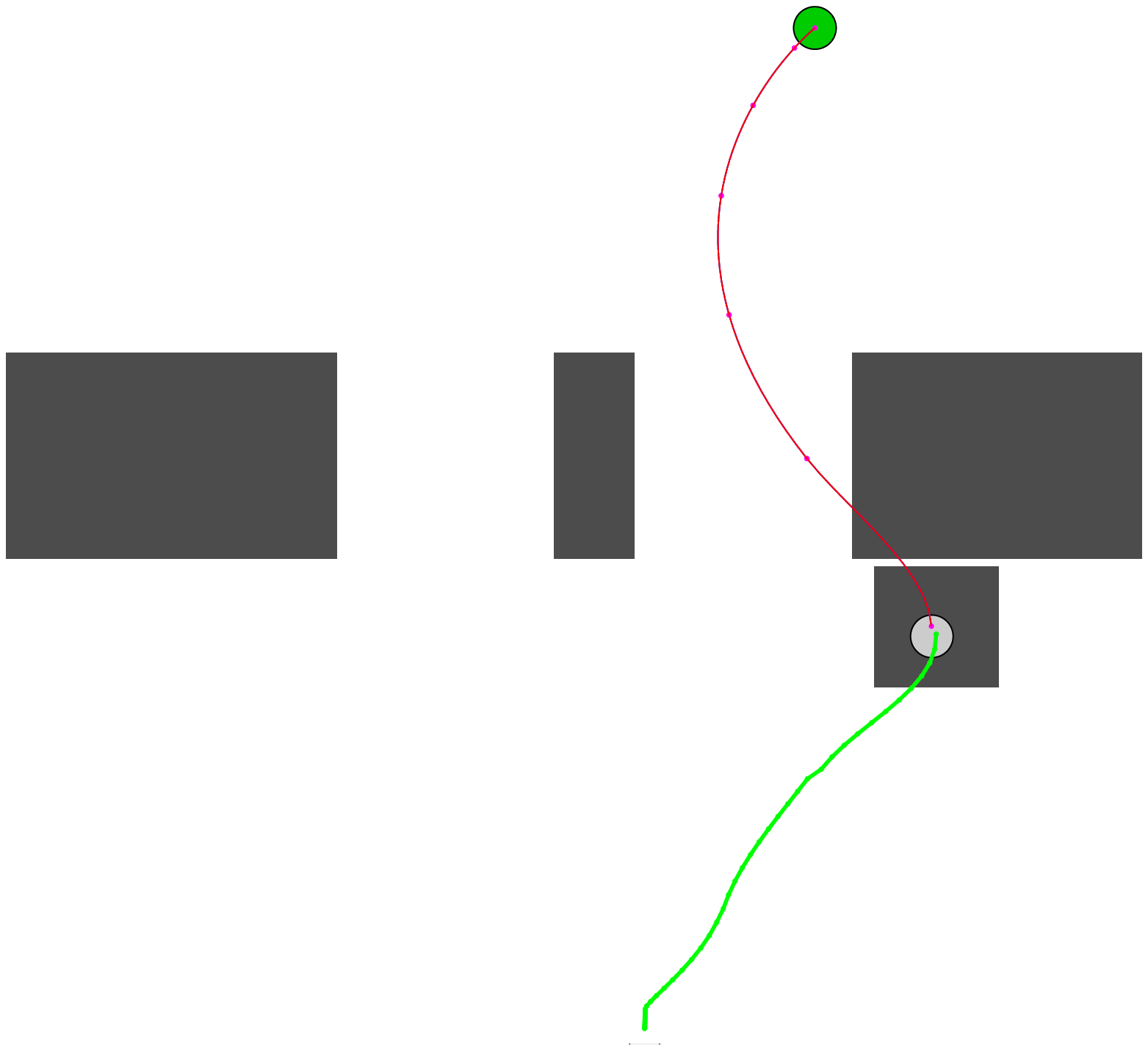}
        \caption{$t=5$}\label{fig:t8}
    \end{subfigure}
	\hfill\vline\hfill
    \begin{subfigure}[t]{0.24\linewidth}
        \includegraphics[trim={60 80 100 90},clip,width=\linewidth]{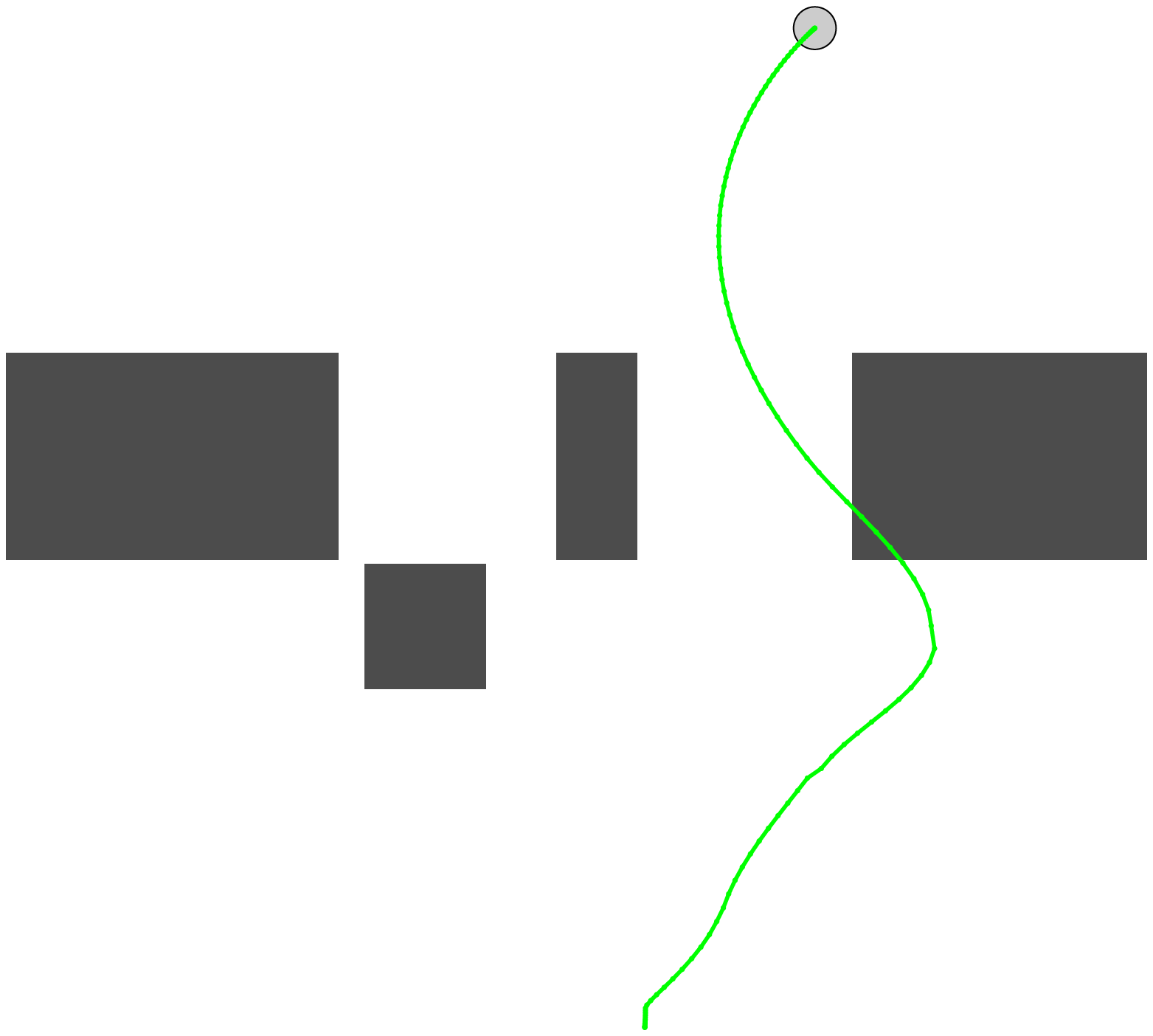}
        \caption{$t=10$}\label{fig:tn}
    \end{subfigure}%
    \caption{\small GPMP-GRAPH in (a) finds the lowest-cost trajectory (red) from the initial graph construction of 2 chains. It optimizes and follows this path at each time step, shown in green. In (b)-(c) the optimized path (red) continues to get pushed laterally due to the moving object, forcing the robot to collide with the obstacle. The final trajectory in (d) shows the robot colliding with the passageway as a result of being unable to successfully plan around the moving obstacle.}
    \label{fig:narrow_passage}
\end{figure*}

The dynamic narrow passageway consists of an obstacle moving back-and-forth with constant velocity in a 1D motion (left to right then right to left) in front of a narrow passageway that contains two entrances as shown in Fig.~\ref{fig:narrow_passage}. As this obstacle moves in front of the passageway it will alternately block one of the entrances. This environment is used to test the algorithms' online planning capabilities in a controlled setting. As we argued previously the ability to switch homotopies will become important in such a scenario.

The results are summarized in Table~\ref{table:NP-methodsResults}. \alg is able to successfully plan 100\% in narrow passage environment, while executing on average 1-2 homotopy switches to explore the both passage entrances. As a result, it also minimizes distance traveled due to this early, prudent homotopy switch. On the other hand, GPMP2 and GPMP-GRAPH are largely unsuccessful at solving this problem as they fail to successfully find collision free paths. While at first, the optimal path from start to goal is obvious for GPMP2, the moving obstacle quickly eliminates this as a viable path, forcing the algorithms' to improvise or fail. They get stuck in local minima as seen in Fig.~\ref{fig:narrow_passage} where the object's lateral motion continues to push the trajectory to the side until finding a collision free path is no longer possible under the given temporal and physical constraints. GPMP-GRAPH's initial solution that spans multiple chains also succumbs to the same issue since it is reduced to a single chain graph after the first time step. On the other hand, Fig.~\ref{fig:toy_breakdown} shows that \alg contends with this local minima problem by utilizing a multi-chain factor graph to explore a path in the other homotopy class once it becomes more viable than the previously desirable homotopy.
It is important to note that the homotopy switches made by GPMP-GRAPH and GPMP2 are a result of the dynamic environment changing the optimized path's \textit{h-signature} rather than an intended algorithmic or strategic homotopy switch.  

\subsection{Dynamic Forest Benchmark}

\begin{figure*}[ht]
    \centering
    \begin{subfigure}[t]{0.24\linewidth}
        \includegraphics[trim={80 80 50 90},clip,width=\linewidth]{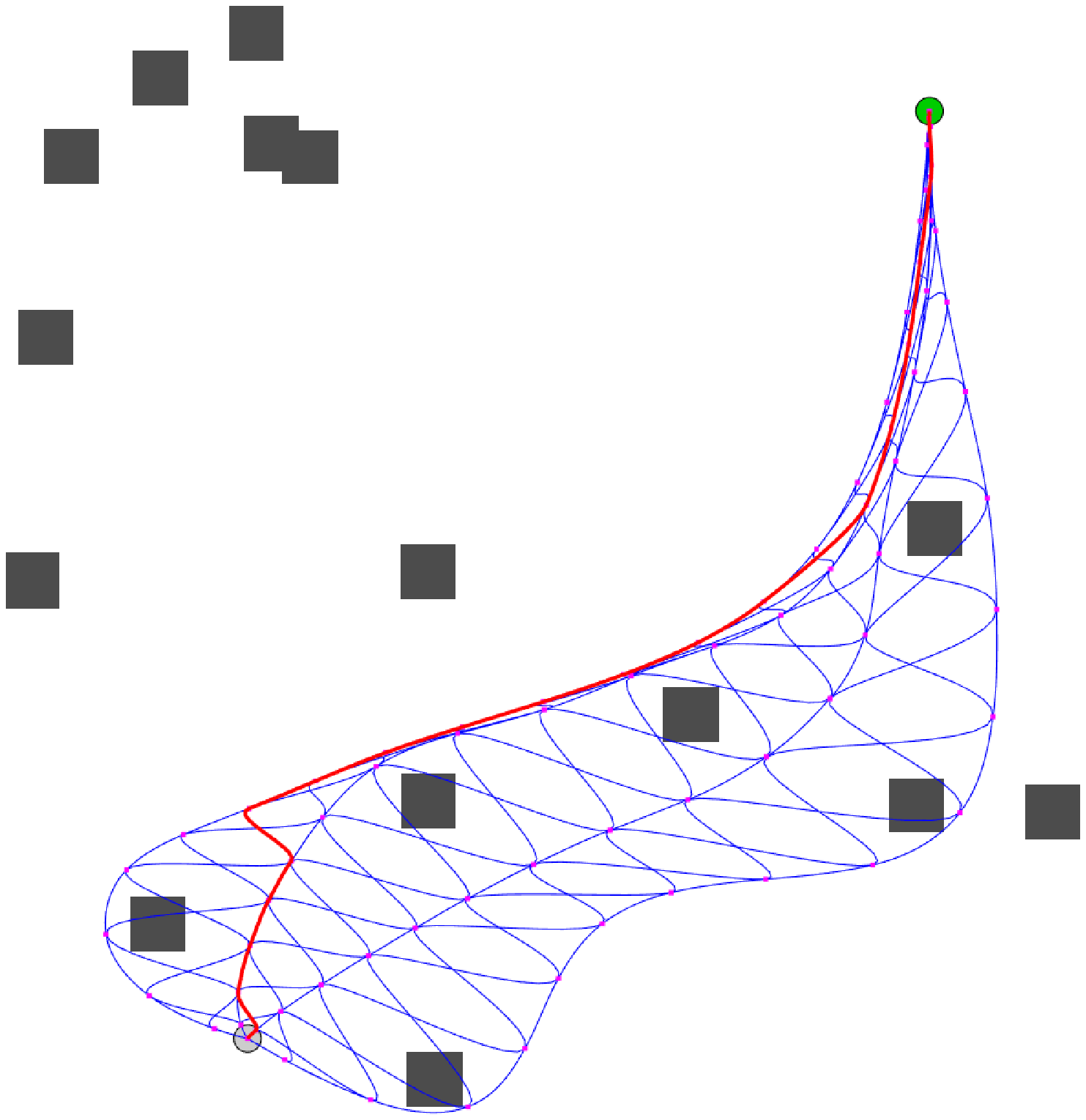}
        \caption{$t = 1$}\label{fig:rf_t1}
    \end{subfigure}%
	\hfill\vline\hfill
    \begin{subfigure}[t]{0.24\linewidth}
        \includegraphics[trim={80 80 50 90},clip,width=\linewidth]{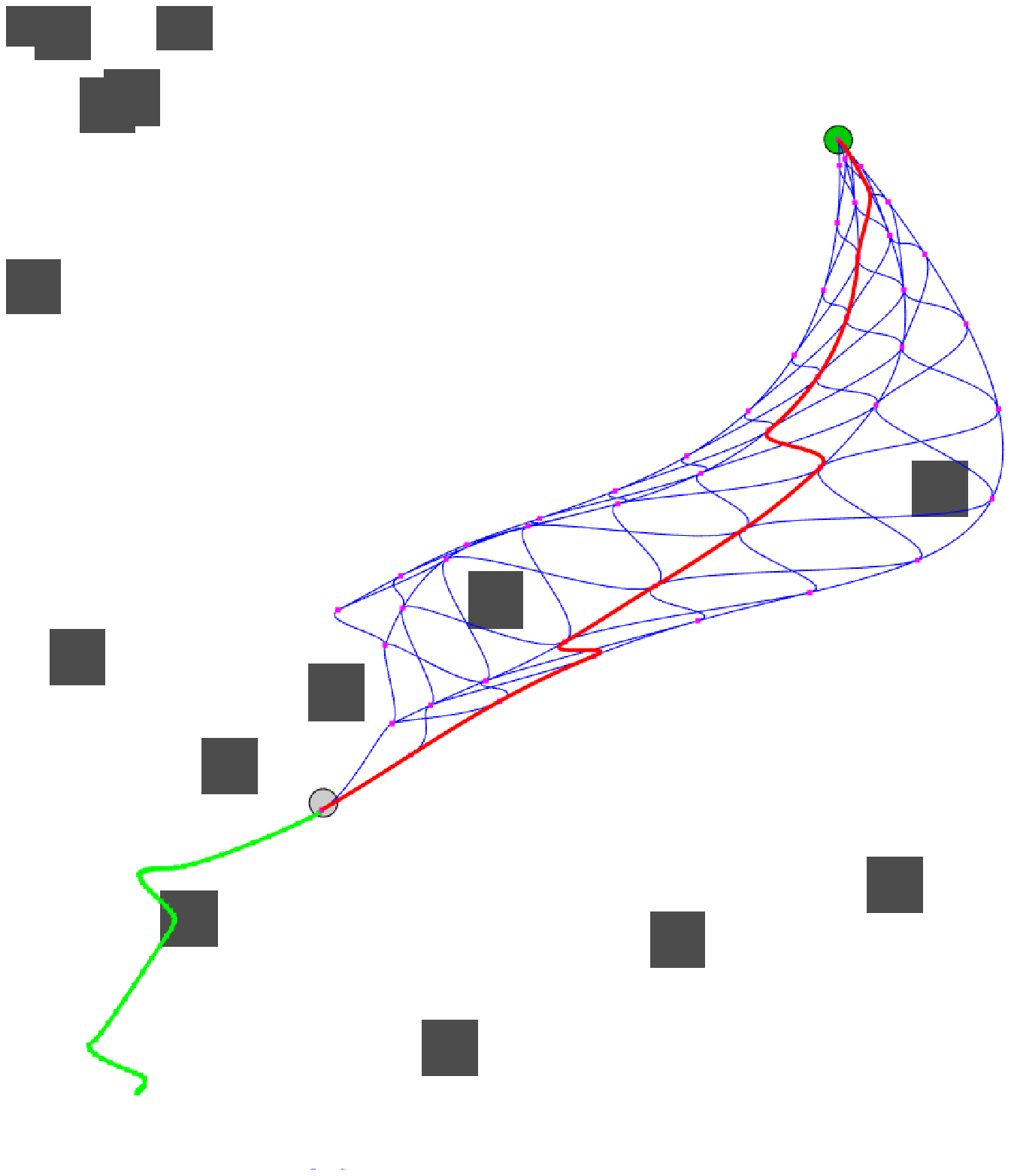}
        \caption{$t = 6$}\label{fig:rf_t2}
    \end{subfigure}%
	\hfill\vline\hfill
    \begin{subfigure}[t]{0.24\linewidth}
        \includegraphics[trim={80 80 50 90},clip,width=\linewidth]{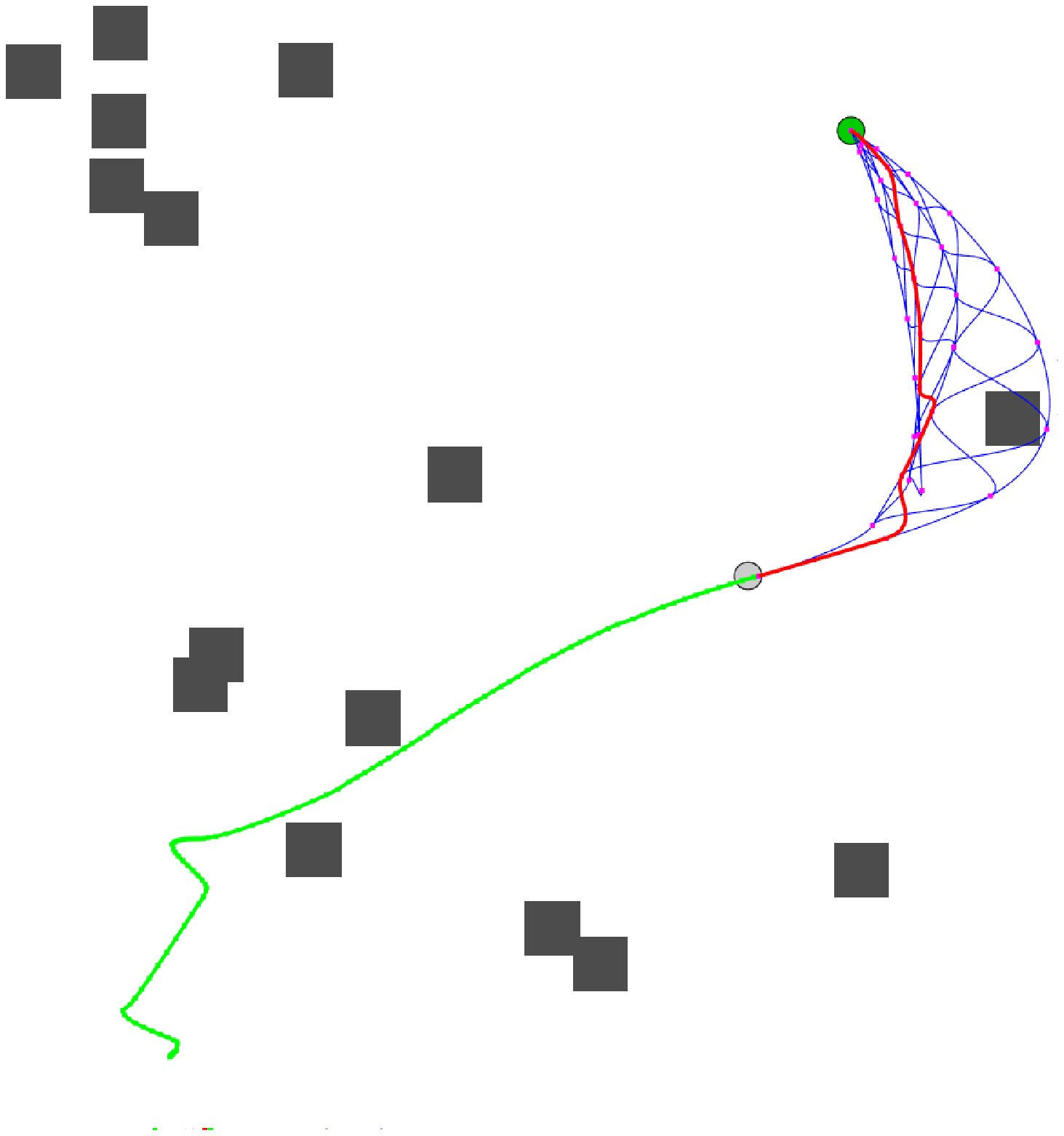}
        \caption{$t = 10$}\label{fig:rf_t3}
    \end{subfigure}%
	\hfill\vline\hfill
    \begin{subfigure}[t]{0.24\linewidth}
        \includegraphics[trim={80 80 50 90},clip,width=\linewidth]{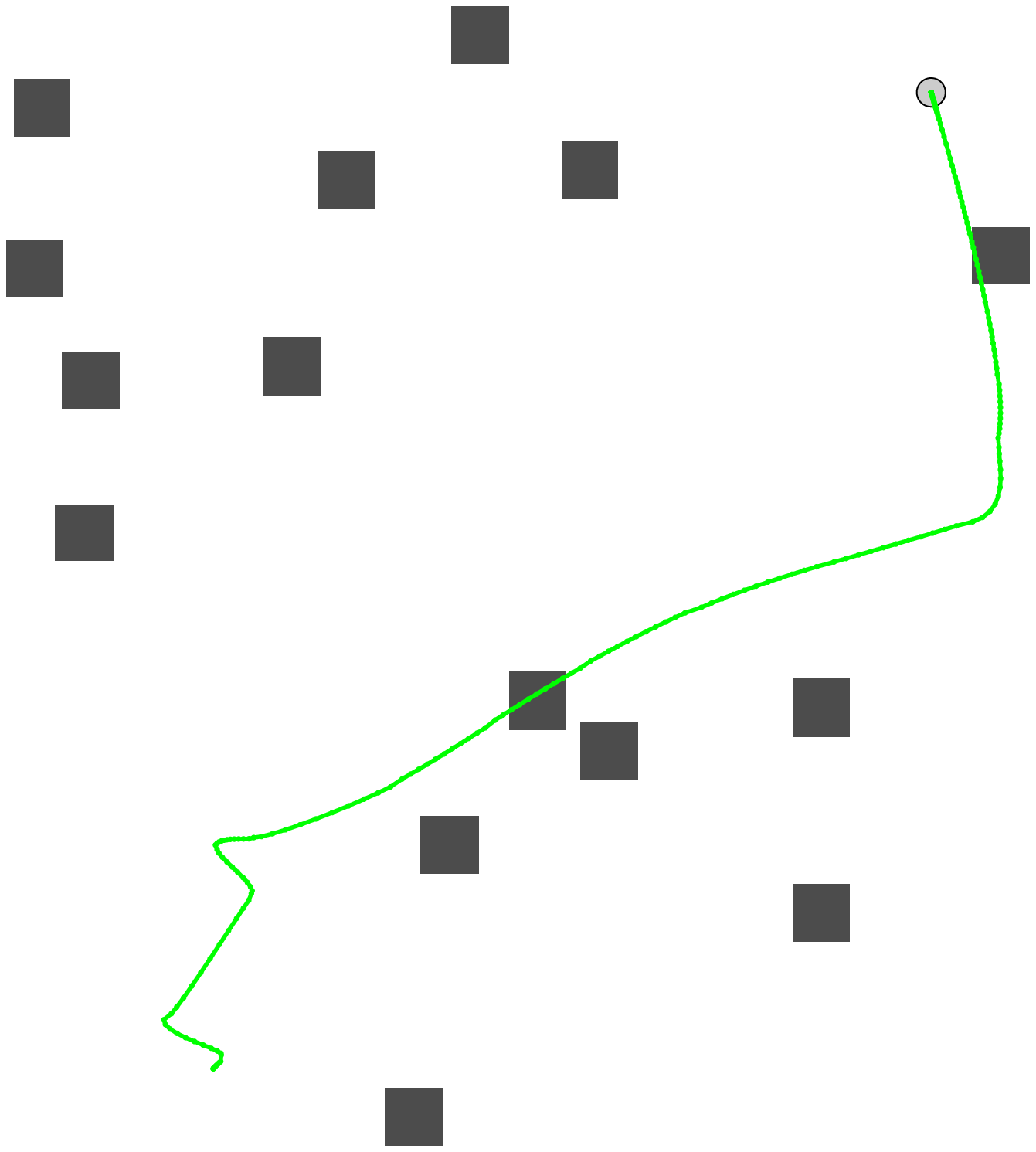}
        \caption{$t = 21$}\label{fig:rf_t4}
    \end{subfigure}%
    \caption{\small \alg successfully returns a collision free trajectory utilizing 4 chains. At each consecutive time-step $t$, the robot (gray circle) prunes away unreachable portion of the graph and reoptimizes the remaining graph until it reaches the goal state (green circle). The path the robot has traveled so far is shown green and the lowest-cost trajectory going forward is shown in red.}
    \label{fig:dynamic_forest_timelapse}
	\vspace{-5mm}
\end{figure*}

The dynamic forest environment consists of numerous objects exhibiting random 2D motion. This environment is simulated given a number of obstacles, their initial state and a random bounded acceleration applied at each time step. The dynamic forest is shown in Fig.~\ref{fig:dynamic_forest}. It is employed to benchmark the ability of the algorithms to contend with a harder replanning problem.

The dynamic forest benchmark results are shown in Table~\ref{table:DF-methodsResults}. Our algorithm generates paths that are more successful and when it fails it spends less time on average in collision. Again, this can be attributed to \alg constantly replanning and reoptimizing the multi-chain factor graph instead of sticking with a single chain factor graph, allowing for online homotopy switching. This process can be seen in Fig.~\ref{fig:dynamic_forest_timelapse} and Fig.~\ref{fig:dynamic_forest}. Since there are many more paths for \alg to consider, it returns a feasible path more often than the two other algorithms.
In Fig.~\ref{fig:dynamic_forest}, the black line represents the trajectory considered best at the previous time step while the red line represents the trajectory considered best at the current time step. Fig.~\ref{fig:switch1} signifies a homotopy switch as the red and black lines cannot be continuously deformed into each other without colliding with obstacles. Fig.~\ref{fig:switch2} shows the robot executing the first step of the red path from Fig.~\ref{fig:switch1} while again considering another homotopy switch.

Note that GPMP2's on average shorter distance traveled can be attributed to the straight line initialization that characterizes the region around the local minima, but this also results in a greater collision intensity and a much lower success rate.

\begin{table}[!t]
	\caption{Benchmark results on dynamic forest dataset.}
	\label{table:DF-methodsResults}
	\begin{center}
	\resizebox{\columnwidth}{!}{
		\begin{tabular}[c]{rccc}
			\toprule
			Metric & \alg & GPMP-GRAPH & GPMP2 \\
			\midrule
			Success Rate (\%) & \textbf{29.23} & 11.92 & 8.46\\
			Collision Intensity (\%) & \textbf{1.9} & 5.1 & 8.3\\
			Distance (m) & 59.05 & 66.52 & \textbf{53.41}\\
			Homotopy Switches & \textbf{5.15} & 4.23 & 3.54\\
			Avg. Comp. Time (s) at $t=0$ & 0.221 & 0.020 & \textbf{0.0197}\\
			Avg. Comp. Time (s) at any $t>0$ & 0.077 & 0.029 & \textbf{0.028}\\ 
			\bottomrule
		\end{tabular}
	}
	\end{center}
	\vspace{-2mm}
\end{table}

\begin{table}[!t]
	\caption{Performance of \alg with varying number of chains on dynamic forest dataset.}
	\label{table:DF_numPathsResults}
	\begin{center}
		\begin{tabular}[c]{rccc}
		\toprule
		Metric & $N_I$=2 & $N_I$=4 & $N_I$=6 \\
		\midrule
		Success Rate (\%) & 22.31 & \textbf{29.23} & 18.85\\
		Collision Intensity (\%) & \textbf{1.9} & \textbf{1.9} & 2.2\\
		Distance (m) & 59.1 & \textbf{59.0} & 60.6\\
		Homotopy Switches & \textbf{5.23} & 5.15 & 4.38\\
		\bottomrule
		\end{tabular}
	\end{center}
	\vspace{-3mm}
\end{table}

We also analyze the performance of \alg on factor graphs with different numbers of chains, $N_I=2$, $N_I=4$, and $N_I=6$. The results in Table~\ref{table:DF_numPathsResults} indicate that there is a sweet spot in the middle (at $N_I=4$) to get the best performance. An example with four chains is visualized in Fig.~\ref{fig:dynamic_forest_timelapse}. The two chain factor graph will have fallen into two of the four homotopy classes that the four chain factor graph will have, thus exhibiting similar metrics. The six chain factor graph exhibits a reduction in performance due to chains collapsing into redundant homotopy classes as a result of $Q_I$ and the number of time steps making the optimization problem too constrained. It is also important to consider the algorithmic computation time when deciding on the number of chains to include in the factor graph as increasing number of chains increases computation time. Ideally the computation time should be smaller than the time length of the trajectory between two time steps ($\delta t$) for \alg to be applicable online, which was indeed the case in our experiments. Table~\ref{table:DF-methodsResults} shows the average computation times for the first iteration at $t=0$ and for any other iteration at $t>0$, where $\delta t = 0.5$s.


\section{Discussion}\label{sec:conc}

In this paper, we introduced \alg, a novel online graph-based trajectory optimization planning algorithm. It builds on recent inference based trajectory optimization techniques to add capability to plan online and dynamically switch homotopies. We found that \alg can leverage knowledge of distinct trajectories across different homotopy classes to dynamically select a low cost solution in light of new information about the environment. In dynamic environments with moving obstacles, previous trajectory-optimization approaches are collision-prone as they become stuck in a local minima. Rather than committing to a low-quality trajectory in a local minima, \alg successfully generates smooth and collision-free trajectories by maintaining and optimizing multiple interconnected solutions to the planning problem.

Our work is currently limited to a finite time horizon setting i.e. the number of discretized time steps we are planning over is fixed. For long range tasks in dynamic environments, occasionally a receding horizon formulation is more beneficial to avoid wasting computation on latter portions of the plan far ahead in the future. Extending our approach to such settings can further improve its capability and applicability. \alg (and GPMP-GRAPH) can sometimes suffer from different trajectories collapsing into the same homotopy class. To contend with this problem it will be beneficial to further investigate principled ways of integrating sampling strategies for new support states such that \alg not only prunes unreachable states, but also add new states at every time step to allow for diversity and exploration. 

\vspace{-1mm}
\bibliographystyle{IEEEtran}
\bibliography{ref}

\end{document}